\documentclass[aps,pre,twocolumn,superscriptaddress]{revtex4-1}
\usepackage{graphicx}
\usepackage{array}
\usepackage{amsfonts}
\usepackage{amssymb,amsmath,multirow,rotate}
\usepackage{mathrsfs}
\usepackage{booktabs}
\usepackage{threeparttable}
\usepackage{multirow}
\usepackage{epsfig}
\usepackage{threeparttable}
\usepackage{chngpage}
\usepackage{latexsym}
\usepackage{dcolumn}
\usepackage{graphics,graphicx,bm,fleqn,epic,eepic,float}
\usepackage{verbatim}
\usepackage{subfigure}
\usepackage{color}

\definecolor{red}{rgb}{1,0,0}

\definecolor{blue}{rgb}{0,0,1}

\begin{document}

\title{Model-free prediction of spatiotemporal dynamical systems with recurrent neural networks: Role of network spectral radius}

\date{\today}

\author{Junjie Jiang}
\affiliation{School of Electrical, Computer and Energy Engineering, Arizona State University, Tempe, AZ 85287, USA}

\author{Ying-Cheng Lai} \email{Ying-Cheng.Lai@asu.edu}
\affiliation{School of Electrical, Computer and Energy Engineering, Arizona State University, Tempe, AZ 85287, USA}
\affiliation{Department of Physics, Arizona State University,
Tempe, Arizona 85287, USA}

\begin{abstract}

	A common difficulty in applications of machine learning is the lack of any general principle for guiding the choices of key parameters of the underlying neural network. Focusing on a class of recurrent neural networks - reservoir computing systems that have recently been exploited for model-free prediction of nonlinear dynamical systems, we uncover a surprising phenomenon: the emergence of an interval in the spectral radius of the neural network in which the prediction error is minimized. In a three-dimensional representation of the error versus time and spectral radius, the interval corresponds to the bottom region of a ``valley.'' Such a valley arises for a variety of spatiotemporal dynamical systems described by nonlinear partial differential equations, regardless of the structure and the edge-weight distribution of the underlying reservoir network. We also find that, while the particular location and size of the valley would depend on the details of the target system to be predicted, the interval tends to be larger for undirected than for directed networks. 
The valley phenomenon can be beneficial to the design of optimal reservoir computing, representing a small step forward in understanding these machine-learning systems.

\end{abstract}
\maketitle

\section{Introduction} \label{sec:intro}

Recent years have witnessed a growing interest in exploiting machine-learning 
algorithms for predicting the state evolution of nonlinear dynamical 
systems~\cite{HSRFG:2015,LBMUCJ:2017,PLHGO:2017,LPHGBO:2017,PWFCHGO:2018,
PHGLO:2018,Carroll:2018,NS:2018,ZP:2018,WYGZS:2019}. Reservoir computing, a 
form of echo state~\cite{Jaeger:2001,MJ:2013} or liquid state~\cite{MNM:2002} 
machines that are fundamentally recurrent neural networks, stands out as a 
viable paradigm for model-free, data based prediction of chaotic 
systems~\cite{JH:2004,PLHGO:2017,LPHGBO:2017,PWFCHGO:2018,PHGLO:2018,ZP:2018}. 
A general reservoir computing scheme consists of an input layer, a reservoir
that is a high-dimensional or networked neural dynamical system, and an output 
layer. The input layer maps the given, low-dimensional time series or 
sequential data into the high-dimensional phase space of the reservoir network,
and the output layer maps the evolution of the high-dimensional dynamical 
variables of the reservoir back into low-dimensional time series as readout. 
During the training phase, the output is compared with the original input 
data from the target system, and parameters of the output layer are tuned to 
minimize the difference. A properly trained reservoir-computing system 
without any input is itself a dynamical system whose evolution from a given 
set of initial conditions represents the prediction of the state evolution 
of the target system from that particular initial-condition set. Since the 
high-dimensional neural network system constituting the reservoir is 
pre-determined and fixed, learning can be accomplished fast with high 
efficiencies and at low cost. Physically, reservoir computing can be realized 
electronically with time-delay autonomous Boolean systems~\cite{HSRFG:2015} 
or implemented using high-speed photonic devices~\cite{LBMUCJ:2017}. 

There are two types of parameters in reservoir computing or in machine learning
in general: a pre-defined, fixed set of parameters and a set of tunable 
parameters whose values are determined through the training or learning process.
For convenience, we call the former {\em free parameters} and the 
latter {\em learning parameters}. An extremely challenging issue in machine
learning is the lack of general rules or criteria for selecting the 
pre-defined parameters. The common practice is mostly a random, brute-force 
type of trial-and-error process to determine the parameter values. Because of 
the vast complexity of the neural network dynamics associated with machine 
learning, to develop any general and systematic approach to choosing the 
pre-defined parameters has remained to be an outstanding problem, with 
little possibility of viable solutions in sight. 

In this paper, we report a general phenomenon associated with reservoir 
computing as applied to model-free and data-based prediction of nonlinear 
dynamical systems, which can be used to guide the choice of the core component 
of the neural computing system: the high-dimensional dynamical backbone
neural network constituting the reservoir. To be as general as possible, we 
assume that the reservoir is described by a complex weighted network. Because
of the large size of the network, a vastly large number of pre-defined 
parameters (and properties) will then need to be determined, such as the 
network topology, the average degree, the network size, and the edge weights, 
and so on, making any systematic selection of these parameters/properties a 
practically impossible task. To make our study feasible, we consider both 
directed and undirected topology and set to fix the network structure,
leaving only the edge weights as the set of free parameters. Even then, the 
possible parameter choices are enormous. Quite surprisingly, we find that, 
in spite of the large number of free parameters, the one that is key to 
success of reservoir computing is the spectral radius of the complex neural
network. In particular, we find that there exists an interval of the values 
of the network spectral radius within which the training error associated with 
reservoir computing is minimized. On a three-dimensional plot of the error 
versus time and spectral radius, a valley-like structure with a flat bottom of 
finite size with near zero error emerges. This means that, regardless of the 
network details, insofar as its spectral radius is chosen from the valley 
region, model-free prediction with reservoir computing can be guaranteed. 
We establish this result through a number of nonlinear dynamical systems
arising from different physical contexts: spatiotemporal systems described 
by the nonlinear Schr\"{o}dinger equation (NLSE), the Kuramoto–Sivashinsky 
equation (KSE), and the one-dimensional complex Ginzburg-Landau equation 
(CGLE). Considering that general phenomena for guiding the choices of 
parameter values are rare in the machine learning literature, our finding 
is encouraging and may stimulate further efforts in searching for common 
principles underlying the working of machine learning not only in reservoir 
computing but also beyond.

\section{Reservoir computing} \label{sec:RC}

\begin{figure} 
\centering
\includegraphics[width=\linewidth]{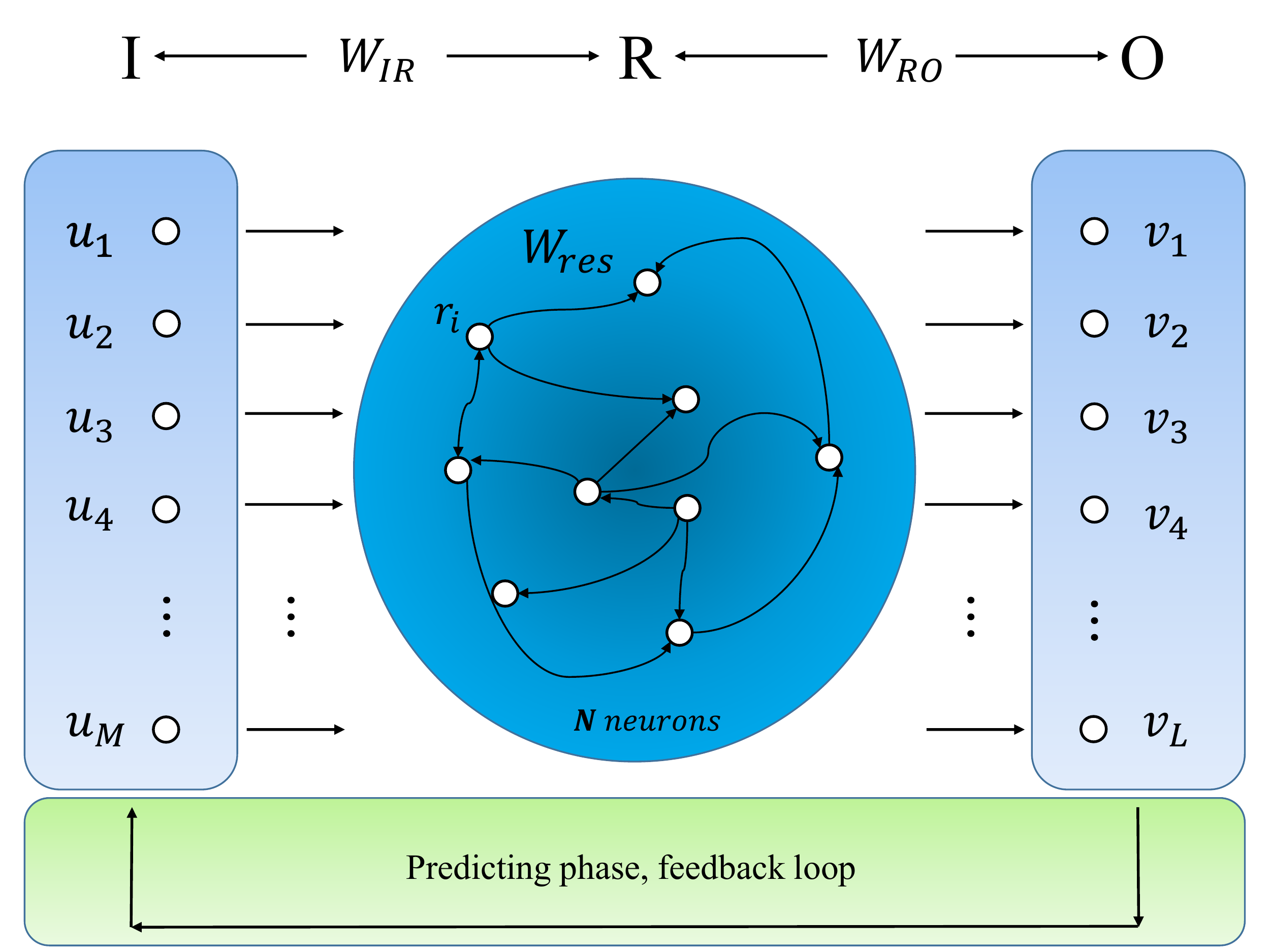}
\caption{{\em Basic structure of reservoir computing.} 
The left blue box represents input the subsystem that maps the $M$-dimensional 
input data to a vector of much higher dimension $N$, where $N \gg M$. The blue 
circle in the middle denotes the reservoir system that can be, for example, a 
complex neural network of $N$ interconnected neurons, whose connection 
structure is characterized by the $N\times N$ weighted matrix 
$\mathbf{W}_{res}$. The dynamical state of the $i^{th}$ neuron in the 
reservoir is $r_i$. The blue box on the right side represents the output 
module that converts the $N$-dimensional state vector of the reservoir 
network into an $L$-dimensional output vector, where $N \gg L$. The mapping 
from the input module to the reservoir is described by the $N\times M$ 
weighted matrix $\mathbf{W}_{IR}$, and that from the reservoir to the output 
module by the $L\times N$ weighted matrix $\mathbf{W}_{RO}$. During the 
training phase, the three blue boxes are activated. In this case, the whole 
computing device is effectively a nonlinear dynamical system with external 
input. In the prediction phase, the external input is cut off and the output 
data are directly fed back to the reservoir (the green box), so the system is 
one without any external driving.}
\label{fig:res_illust}
\end{figure}

There are two major types of reservoir computing systems: echo state networks 
(ESNs)~\cite{Jaeger:2001} and liquid state machines (LSMs)~\cite{MNM:2002}. 
The architecture of an ESN is one that is associated with supervised learning 
underlying recurrent neural networks (RNNs). The basic principle of ESNs 
is to drive a large neural network of a random or complex topology - the 
reservoir network, with the input signal. Each neuron in the network generates 
a nonlinear response signal. Linearly combining all the response signals 
with a set of trainable parameters yields the output signal. As for ESNs, 
an LSM is also a random or complex network of neurons with the difference 
being that each neuron receives not only an external input signal but also 
signals from other neurons in the network. The networked system is thus 
effectively a spatiotemporal nonlinear dynamical system, where trainable, 
linearly discriminant units are used to map the spatiotemporal patterns of 
the network into proper output signals. Structure-wise, LSMs are more 
complicated than ESNs. 

For simplicity, we focus on ESNs. A schematic illustration of a typical ESN is 
shown in Fig.~\ref{fig:res_illust}, where the reservoir computing machine 
consists of three components: (i) an input subsystem that maps the 
low-dimensional (say $M$) input signal into a (high) $N$-dimensional signal 
through the weighted $N\times M$ matrix $\mathbf{W}_{IR}$, (ii) the reservoir 
network of $N$ neurons characterized by $\mathbf{W}_{res}$, a weighted network 
matrix of dimension $N\times N$, and (iii) an output subsystem that converts 
the $N$-dimensional signal from the reservoir network into an $L$-dimensional 
signal through the output weighted matrix $\mathbf{W}_{RO}$, where 
$L \sim M \ll N$. In Fig.~\ref{fig:res_illust}, the three components are
denoted as I, R and O, respectively. 

The working of an ESN can be described, as follows. As shown in 
Fig.~\ref{fig:res_illust}, the training phase is represented by the blue 
blocks. The input multidimensional data has the dimension $M\times N_t$, 
where $M$ is the dimension of the input data vector $\mathbf{u}(t)$ at time 
$t$ and $N_t$ is the number of time steps used in the training phase: 
$t=0,dt,2dt,\ldots,(N_t-1)dt$. The input data vector to the reservoir network 
is $\mathbf{W}_{IR}\cdot\mathbf{u}(t)$. The values of the elements in 
$\mathbf{W}_{IR}$ are obtained from a uniform distribution in 
$[-\alpha,\alpha]$. Every neuron in the reservoir receives one component of 
the input data vector to the reservoir. Typically, the reservoir is a large, 
sparse, directed or undirected random network with average degree $k$, which 
is described by a weighted adjacency matrix $\mathbf{W}_{res}$, whose largest 
absolute eigenvalue is the network spectral radius $\rho$. For a given value 
of $\rho$, we choose the values of all the elements of $\mathbf{W}_{res}$ 
randomly from a uniform distribution and rescale all the values so that its 
largest eigenvalue is $\rho$. The state of the whole reservoir at time $t$ 
is a $N$-dimensional vector $\mathbf{r}(t)$, where each dimension represents 
the dynamical state of an individual node. For the $i$th node, its state is 
denoted by $r_i(t)$. The initial state of the reservoir is 
$\mathbf{r}(0) = {\bf 0}$. The state of reservoir is updated at 
every time step $dt$ according to
\begin{equation} \label{eq:res_evol}
\mathbf{r}(t+dt)=f(\mathbf{W}_{res}\cdot r(t)+
\mathbf{W}_{IR}\cdot\mathbf{u}(t)),
\end{equation}
where $f$ is a function that activates every element in the reservoir state 
vector, a typical choice of which is the $\mbox{tanh}$ function. After 
evolving Eq.~(\ref{eq:res_evol}) for all the time steps in $\mathbf{u}$, we get
an $N\times (N_{t}+1)$-dimensional matrix of the reservoir state $\mathbf{r}$. 
We disregard the first $S$ steps of the reservoir as transients. Since the 
activation function $\mbox{tanh}$ is odd, it is necessary~\cite{PHGLO:2018}
to normalize the vector $\mathbf{r}$ by taking the squares of its even elements.
This leads to a new state vector $\mathbf{r'}$. All the training is done with 
respect to the normalized reservoir state vector $\mathbf{r'}$ and the output 
vector $\mathbf{v}$, which updates the output matrix $\mathbf{W}_{RO}$. 
The training phase is completed when the output $L\times N$ matrix 
$\mathbf{W}_{RO}$ has been determined.

In the prediction phase, all the blue and green blocks in 
Fig.~\ref{fig:res_illust} are activated. The only difference from the learning
phase is that the input data $\mathbf{u}(t+dt)$ is replaced by $\mathbf{v}(t)$. 
More specifically, to predict the dynamical states of the underlying system, 
we set $\mathbf{v}$ to be the input data set $\mathbf{u}$ at the next time 
step: $\mathbf{v}(t)=\mathbf{u}(t+dt)$. The output matrix $\mathbf{W}_{RO}$ 
can be calculated using the regression scheme that minimizes the loss function: 
\begin{equation}
	\mathcal{L}=\sum_{t=d+1}^{N_{t}}\Vert\mathbf{v}(t)-\mathbf{W}_{RO}\mathbf{r'}(t)\Vert+\Gamma\Vert\mathbf{W}_{RO}\Vert^2,  
\end{equation}   
where $\Vert\mathbf{W}_{RO}\Vert^2$ is the sum of squared elements of 
$\mathbf{W}_{RO}$. The parameter $\Gamma$ is a small positive regularization 
constant introduced for preventing over-fitting by imposing penalty on large 
values of the fitting parameters. The regularized regression can be described 
as
\begin{equation}
	\mathbf{W}_{RO}=\mathbf{v}\cdot\mathbf{r'}^T\cdot\mathbf{r'}\cdot(\mathbf{r'\cdot r'}^T+\Gamma\mathbf{I})^{-1}, 
\end{equation}

There are two types of strategies to set the initial state of the reservoir
network. One can simply continue from the training phase to predict the 
system, i.e., one continues to use the reservoir state at the final time step 
of the training phase for prediction, entailing a ``warm start'' of the 
prediction phase. Alternatively, one can start the prediction from a different 
data set, where the initial reservoir state is an $N$-dimensional zero vector. 
For a number of time steps at the beginning of the prediction phase, one uses 
the true state $\mathbf{u}$ of the system in Eq.~(\ref{eq:res_evol}) to drive 
the reservoir to a functioning state. This is essentially a ``cold state'' 
strategy.

After a ``warm'' or a ``cold'' start, the dynamical state of the reservoir 
has been activated. With the $\mathbf{W}_{RO}$ matrix determined during the 
training phase, the output of the reservoir is given by
\begin{equation}
	\mathbf{v}(t-dt)=\mathbf{W}_{RO}\cdot \mathbf{r'}(t), 
\end{equation}
where $\mathbf{r'}(t)$ has been updated from $\mathbf{r}(t)$ with the 
elements in the even rows squared. After obtaining $\mathbf{v}(t-dt)$, one 
replaces $\mathbf{u}(t)$ by $\mathbf{v}(t-dt)$ and the reservoir system can 
produce the predicting time series continuously. This feedback process is 
illustrated by the green block in Fig.~\ref{fig:res_illust}.

In a recent work~\cite{PHGLO:2018}, ESNs have been applied to predicting 
the dynamical state of the spatially extended KSE in the chaotic 
regime. It was demonstrated that, with properly chosen parameters, an 
ESN can predict the dynamical states of the KSE in the entire spatial
domain for several Lyapunov time.

\section{Fundamental role of spectral radius of reservoir network in 
predicting spatiotemporal dynamical systems} \label{sec:rho}

We demonstrate that the spectral radius of the reservoir complex network 
plays a fundamental role in achieving successful prediction. We substantiate 
this finding through a number of spatiotemporal dynamical systems arising 
from physics: the NLSE, the KSE, and the CGLE.

\subsection{Nonlinear Schr\"{o}dinger equation} \label{subsec:NLSE}

The NLSE has been a paradigm to study nonlinear wave propagation in fields such
as optics and hydrodynamics~\cite{DDEG:2014}. Among the analytic solutions of  
the NLSE are physically significant phenomena such as ``breathers,'' 
``solitons,'' or rogue waves in a finite background that has been 
experimentally observed in nonlinear fiber 
optics~\cite{SRKJ:2007,DGDKA:2009,KFFMDGAD:2010}. To be concrete, we 
investigate the feasibility of exploiting reservoir computing for predicting 
the finite-background soliton solutions in the NLSE, with a particular eye 
towards elucidating the role of network spectral radius in the prediction.

\begin{figure*} 
\centering
\includegraphics[width=0.9\linewidth]{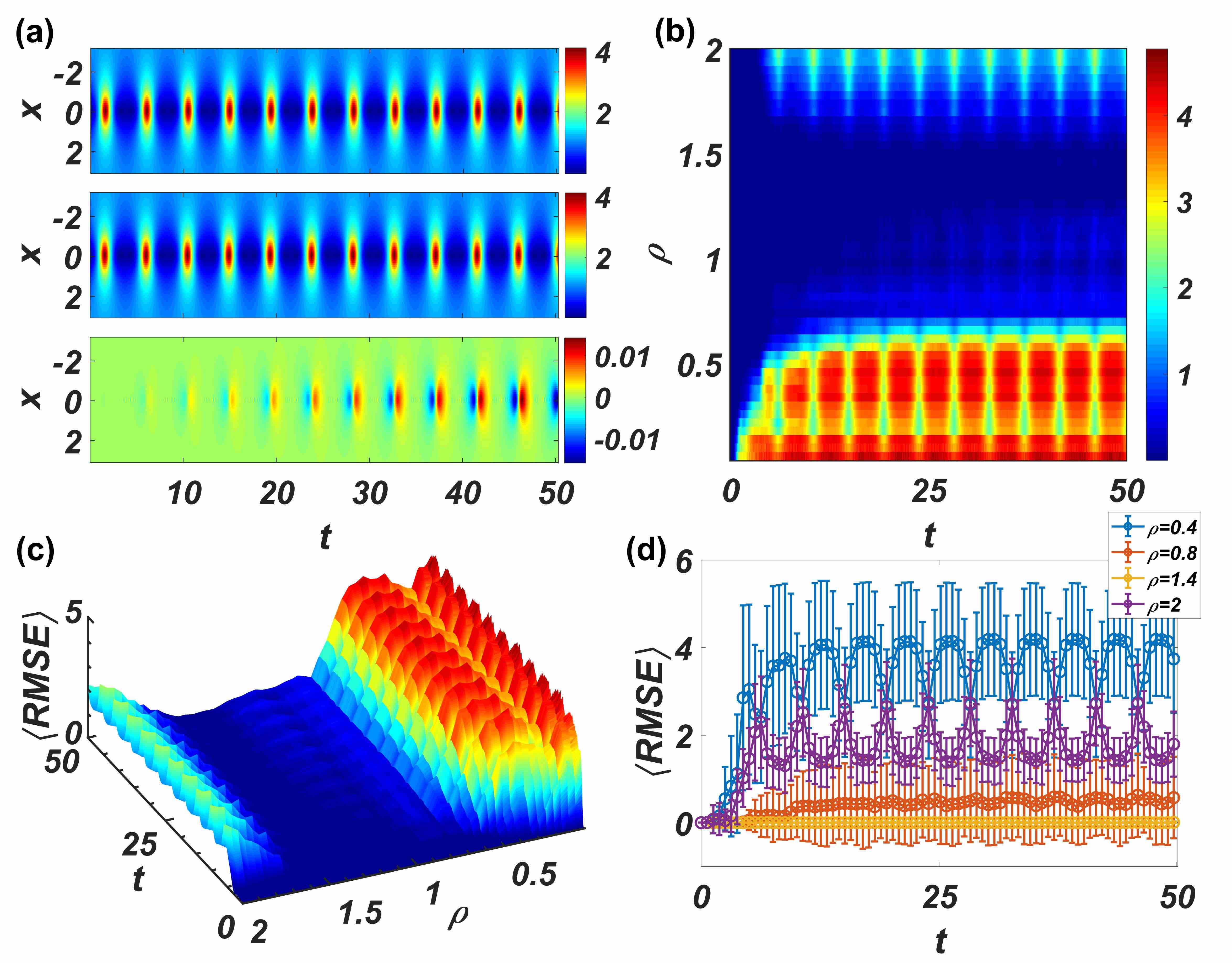}
\caption{ {\em Using reservoir computing to predict Akhmediev breathers and 
the emergence of an optimal interval in the spectral radius}. (a) For spectral 
radius $\rho=1.5$, a successful case of prediction of Akhmediev breathers, 
where the top panel shows the time evolution of the true solution, the middle 
panel displays the predicted solution from reservoir computing after training, 
and the bottom panel depicts the difference between the true and predicted 
solutions. The color bar indicates the scale of the spatiotemporal wave
magnitude $|\psi(x,t)|$. (b) For $0 < \rho \le 2$ (the ordinate), time 
evolution of the ensemble averaged RMSE, denoted as $\langle\mbox{RMSE}\rangle$
where, for each fixed value of $\rho$, $100$ random realizations of the 
reservoir system are used to calculate the average and the color bar indicates 
the scale of the $\langle \mbox{RMSE}\rangle$ values. (c) A three-dimensional 
view of (b). The emergence of a valley interval in $\rho$ with minimized 
prediction error can be seen unequivocally from (b,c). (d) Detailed time 
evolution of $\langle \mbox{RMSE}\rangle$ for four specific values of $\rho$ 
with standard deviation. (More quantitative details can be found in 
Appendix Secs.~\ref{app_sec:SD} and \ref{app_sec:LTP}.) Other parameters of the
reservoir-computing system are $\alpha=1$, $N=4992$, $k=3$, $M=L=64$, 
$N_{t}=8010$, $S=10$, and $\Gamma=1\times 10^{-4}$.}
\label{fig:res_pre_ABs}
\end{figure*}

The finite-background soliton solutions include Akhmediev breathers, 
Kuznetsov-Ma solitons, and Peregrine solitons. We consider analytic solutions 
of the NLSE representing Akhmediev breathers and Kuznetsov-Ma solitons. The 
dimensionless NLSE reads
\begin{align} \label{eq:NLSE_dimensionless}
\centering
i\frac{\partial\psi}{\partial x}+\frac{1}{2}\frac{\partial^2\psi}{\partial t^2}+|\psi|^2\psi=0,
\end{align}    
where the envelope $\psi(x,t)$ is a function of propagation distance $x$ 
and co-moving time $t$. The analytic solution of the NLSE describing general 
modulation instabilities in optics was first obtained in Ref.~\cite{AK:1986}, 
which is given by
\begin{align} \label{eq:NLSE_solution_1}
\centering
\psi(x,t)=e^{ix}\bigg[1+\frac{2(1-2a)\cosh(bx)+ib\sinh(bx)}{\sqrt{a}\cos(\omega t)-\cosh(bx)}\bigg],
\end{align}
where $b=\sqrt{8a(1-2a)}$, $\omega=\sqrt{2(1-2a)}$, and the positive parameter
$a$ determines the physical properties of the solution. For example, for 
$a=0.25$, the solution corresponds to Akhmediev breathers and, for $a=0.7$,
Kuznetsov-Ma solitons arise. To generate the true data from Akhmediev 
breathers, we set $x\in[-\pi,\pi]$ and discretize the space into $64$ lattice
points. The spatial step size is $dx=2\pi/63$ and the time step is 
$dt=\pi/100$. To generate the true Kuznetsov-Ma solitons data, we set 
$T\in[-\pi,\pi]$ and employ the same spatial discretization scheme. 
Note that, physically, the Akhmediev breathers and Kuznetsov-Ma solitons are 
qualitatively similar through an exchange of the time and space variables.

We also test a more complicated type of wave patterns, those generated by 
the collision of two solitons~\cite{ASA:2009,FKM:2013}, where the 
corresponding solution can be obtained as the nonlinear superposition of two 
first-order Akhmediev breathers for $0<a_1,a_2<0.5$:
\begin{align} \label{eq:AB_Collision}
\centering
\psi_{12}(x,t)=\psi_0+\frac{2(l_1^*-l_1)s_1r_1^*}{|r_1|^2+|s_1|^2}+\frac{2(l_2^*-l_2)s_{12}r_{12}^*}{|r_{12}|^2+|s_{12}|^2},
\end{align}
where $l_1=i\sqrt{2a_1}$, $l_2=i\sqrt{2a_2}$, and $*$ represents the complex 
conjugate. (A complete expression of solution is presented in 
Appendix~\ref{app_sec:NLSE_solution}.) To generate the data for the soliton 
collision wave pattern, we again set $x\in[-\pi, \pi]$ and discretize the space into $64$ points. The time 
step is $dt=\pi/40$. For illustrative purpose, we consider two parameter 
settings: ($a_1=0.14$, $a_2=0.34$) and ($a_1=0.42$, $a_2=0.18$).  

\begin{figure*} [ht!] 
\centering
\includegraphics[width=\linewidth]{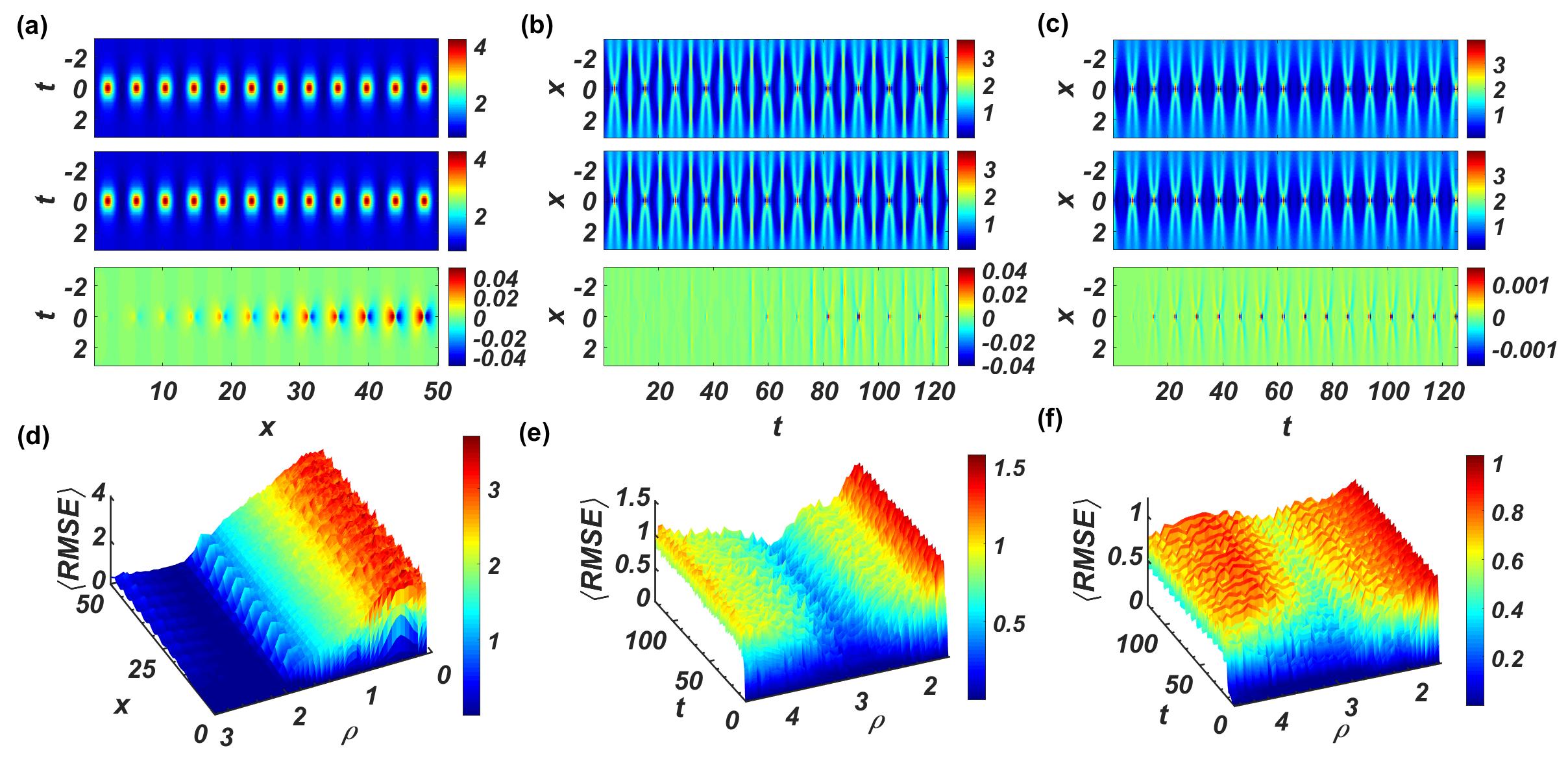}
\caption{ {\em Emergence of a valley in the prediction error versus the 
spectral radius of the reservoir network for predicting Kuznetsov-Ma solitons
and soliton collision in the NLSE}. (a-c) True (upper) and predicted (middle) 
wave patterns as well as the difference (lower) in wave function for 
Kuznetsov-Ma solitons, soliton collision for ($a_1=0.14$, $a_2=0.34$) and 
($a_1=0.42$, $a_2=0.18$), respectively. (d-f) The corresponding time evolution 
of the ensemble averaged prediction error $\langle \mbox{RMSE}\rangle$ for 
systematically varying $\rho$ values. For each fixed $\rho$ value,  $100$ 
random reservoir systems are used to calculate the average error and the
color bar indicates the scale of the $\langle \mbox{RMSE}\rangle$ value. 
For predicting the Kuznetsov-Ma solitons, if the value of $\rho$ is chosen
from the valley, then all realizations lead to near zero errors (d). However,
for the case of predicting soliton collision, only about half of the 
realizations yield near zero errors even when the value of $\rho$ is chosen
from the valley (outside the valley, large errors arise for nearly all 
realizations). Other parameter values are $N=4992$, $k=3$, $M=L=64$, 
$N_{t}=8010$, $S=10$, $\Gamma=1\times 10^{-4}$, $\alpha = 1.0$ for 
panels (a,d), and $\alpha=3.0$ for panels (b,c,e,f). Information about the 
standard deviation of $\langle \mbox{RMSE}\rangle$ can be found in 
Appendix Secs.~\ref{app_sec:SD} and \ref{app_sec:LTP}.}
\label{fig:res_pre_KM_ABsC}
\end{figure*}

Figure~\ref{fig:res_pre_ABs} shows the results of using reservoir computing 
to predict the spatiotemporal evolution of Akhmediev breathers for $a=0.25$.
The dimension of the input data is 64, so is that of the output data. The 
number of nodes (neurons) in the reservoir network is chosen to be 
$N = 4992 =64\times 78$, so every dimension of the input data is connected 
with $78$ neurons in the reservoir. We choose $N_t=8010$ with the transient 
time $\tau=10$, so the training phase contains approximately $56$ solitons. 
In the prediction phase, we choose the strategy of ``warm start'' to initiate 
the dynamical evolution of the reservoir neural network. 
Figure~\ref{fig:res_pre_ABs}(a) shows the results of predicting 11 Akhmediev 
breathers over $1600$ time steps (corresponding to $t \approx 50$). We see
that the occurrence in time of approximately five solitons can be predicted
with relatively small error. 

To search for any possible general rule that can lead to prediction success 
as exemplified in Fig.~\ref{fig:res_pre_ABs}(a), we systematically vary the
value of the spectral radius $\rho$. Extensive tests reveal a remarkable
phenomenon: the emergence of an optimal interval of the $\rho$ values in which 
the prediction error is minimized, as shown in Fig.~\ref{fig:res_pre_ABs}(b),
where the time evolution of the ensemble average of the root mean square 
error ($\langle \mbox{RMSE}\rangle$) between true and predicted solutions for 
$0 < \rho \le 2$ is displayed. For each fixed $\rho$ value, we generate $100$ 
reservoir systems with random weights for $\mathbf{W}_{IR}$, where the 
reservoir network has a random topology with randomly distributed edge
weights (so $\mathbf{W}_{res}$ is effectively a random matrix), and calculate
$\langle \mbox{RMSE}\rangle$. Figure~\ref{fig:res_pre_ABs}(c) is a 
three-dimensional representation of Fig.~\ref{fig:res_pre_ABs}(b), where the 
existence of the optimal $\rho$ interval with minimum error can be identified: 
$\rho \in [0.8,1.6]$. In fact, for $\rho < 0.7$, the values of the error 
$\langle \mbox{RMSE}\rangle$ are dramatically large in comparison with those 
in the valley. As the value of $\rho$ is increased from about 1.6, the error 
value grows approximately linearly. Figure~\ref{fig:res_pre_ABs}(d) shows the 
detailed time evolution of the error $\langle \mbox{RMSE}\rangle$ (with 
standard deviation) for $\rho=0.4, 0.8, 1.4, 2$, where the case of $\rho=1.4$ 
(in the valley) has near zero values of $\langle \mbox{RMSE}\rangle$ as well 
as near zero standard deviation. The results in Fig.~\ref{fig:res_pre_ABs} 
thus indicate that, insofar as the spectral radius of the random reservoir 
network is chosen from the valley, the reservoir system performs well for 
predicting the Akhmediev breathers, regardless of the network structure
and edge-weight distribution. (More detailed information about the behavior 
of RMSE in this case can be found in Appendix Secs.~\ref{app_sec:SD} and 
\ref{app_sec:LTP}.) 

The existence of an optimal interval in the spectral radius of the reservoir
network also holds for Kuznetsov-Ma solitons and colliding solitons.
Specifically, Figs.~\ref{fig:res_pre_KM_ABsC}(a,d) show that a properly 
designed reservoir computing system can predict the solutions of the NLSE 
in the regime of Kuznetsov-Ma solitons, where (d) reveals that the ensemble 
averaged error $\langle \mbox{RMSE}\rangle$ is minimized for 
$\rho \in [1.98,2.34]$ - the valley. [Note that the $\rho$ value of the 
reservoir network in (a) is one with the minimum error in (d).] As the value 
of $\rho$ is decreased from the valley interval, the error 
$\langle \mbox{RMSE}\rangle$ increase rapidly. However, 
$\langle \mbox{RMSE}\rangle$ tends to increase slowly when the value of 
$\rho$ is larger than the valley interval. While the behaviors are similar 
to those in the prediction of Akhmediev breathers, the locations of the
valley interval for the two cases are different, where for the regime of 
Kuznetsov-Ma solitons, the valley occurs in a relatively larger interval
of $\rho$ values.

\begin{figure*} [ht!] 
\centering
\includegraphics[width=0.9\linewidth]{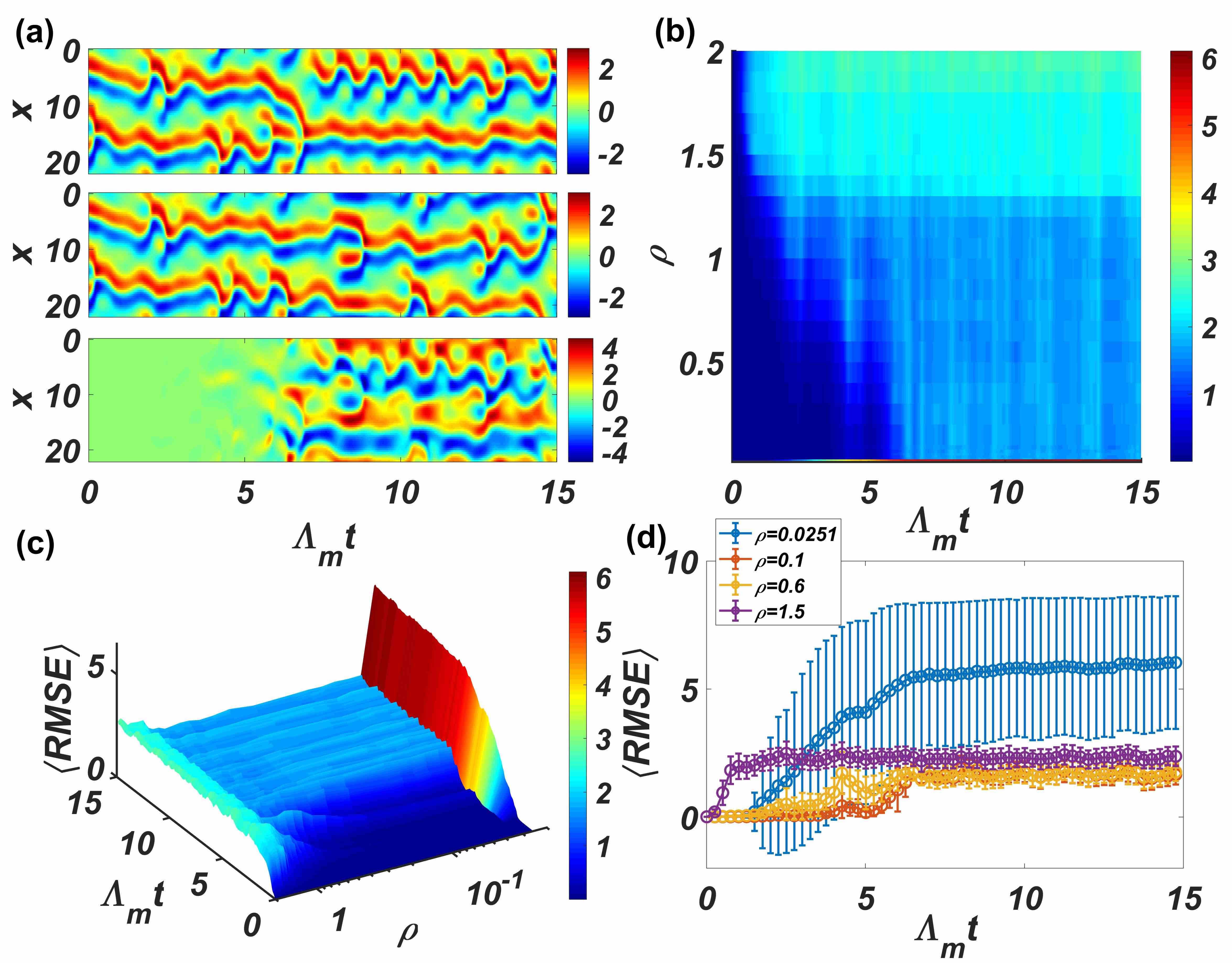}
\caption{ {\em Emergence of a valley interval in the spectral radius of the 
reservoir network with minimum error for predicting spatiotemporal chaotic 
solutions of the KSE}. (a) An example of successful prediction for $\rho=0.1$:
top panel - true spatiotemporal evolution of a typical chaotic solution of 
the KSE; middle panel - predicted spatiotemporal evolution; lower panel - the 
difference between the true and the predicted solutions (true minus predicted).
The color bar indicates the scale of $u(x,t)$. (b) Time evolution of ensemble 
averaged RMSE (over 100 random realizations of the reservoir system) for  
systematically varying $\rho$ values in the interval $[0.0,2.0]$, where the
color bar indicates the scale of $\langle \mbox{RMSE}\rangle$. Note that the 
evolution time is represented as $\Lambda_m t$, where $\Lambda_m \approx 0.05$ 
is the maximum Lyapunov exponent of the chaotic solution and one unit of 
$\Lambda_m t$ corresponds to one Lyapunov time. (c) A three-dimensional 
representation of $\langle \mbox{RMSE}\rangle$ in the $(\rho,\Lambda_m t)$ 
plane, revealing the emergence of a valley interval in $\rho$ in which 
$\langle \mbox{RMSE}\rangle$ is minimized. (d) Time evolution of 
$\langle \mbox{RMSE}\rangle$ for four representative value of $\rho$ with 
standard deviation (more details are in Appendix~\ref{app_sec:SD}). Among 
the four cases, the best prediction result is achieved for $\rho = 0.1$, 
where the chaotic solution can be predicted with near zero error for about 
five Lyapunov time. Other parameters of the reservoir computing system are 
$\alpha=1$, $N=4992$, $k=3$, $M=L=64$, $N_{t}=70010$, $S=10$, and 
$\Gamma=1\times 10^{-4}$.}
\label{fig:res_pre_KS}
\end{figure*}

The results from predicting the patterns of colliding solitons are shown
in Figs.~\ref{fig:res_pre_KM_ABsC}(b,c). While reservoir computing is 
able to predict wave patterns, in this case there is no guarantee that, 
for every random reservoir network with $\rho$ value taken from the valley 
indicated in Figs.~\ref{fig:res_pre_KM_ABsC}(e,f), prediction can be
successful. For example, for $\rho=2.88$, about $50\%$ of the cases 
can yield good prediction result. However, if the value of $\rho$ is not
chosen from the valley interval, reservoir computing fails to make any
meaningful prediction. A similar behavior has been found for a different
case of colliding solitons. We note that, when predicting Akhmediev breathers 
and Kuznetsov-Ma solitons, successful prediction can be achieved for any 
random network whose value of the spectral radius lies in the valley. 
The case of colliding solitons in the NLSE is thus more unpredictable than 
Akhmediev breathers and Kuznetsov-Ma solitons. Nonetheless, in spite of 
the difficulty, the existence of a valley interval leading to optimal 
prediction performance also holds for the case of colliding solitons.

The solution of soliton collision in the NLSE
represents a difficult case where, as shown in Fig.~\ref{fig:res_pre_KM_ABsC},
even for the optimal value of the spectral radius, only about 50 out of 100 
ensemble realizations lead to acceptable prediction results in terms of both 
accuracy and time. The main reason is that the process of soliton collision 
necessarily involves a possible change in the ``climate'' of the dynamical 
state, as the two solitons can bounce back from each other or 
merge~\cite{PKMF:2003}. In the case of bouncing back, the feature or climate 
of the dynamical state of the system remains unchanged before and after the 
collision. In this case, the reservoir computing system is able to make 
accurate predictions. In the latter case of merging, the state climate has 
changed completely before and after the collision, rendering inaccurate 
predictions, as the neural network was mostly trained in the presence of 
two solitons. 

\subsection{Predicting spatiotemporal chaotic solutions of Kuramoto-Sivashinsky equation}

\begin{figure*} [ht!] 
\centering
\includegraphics[width=0.9\linewidth]{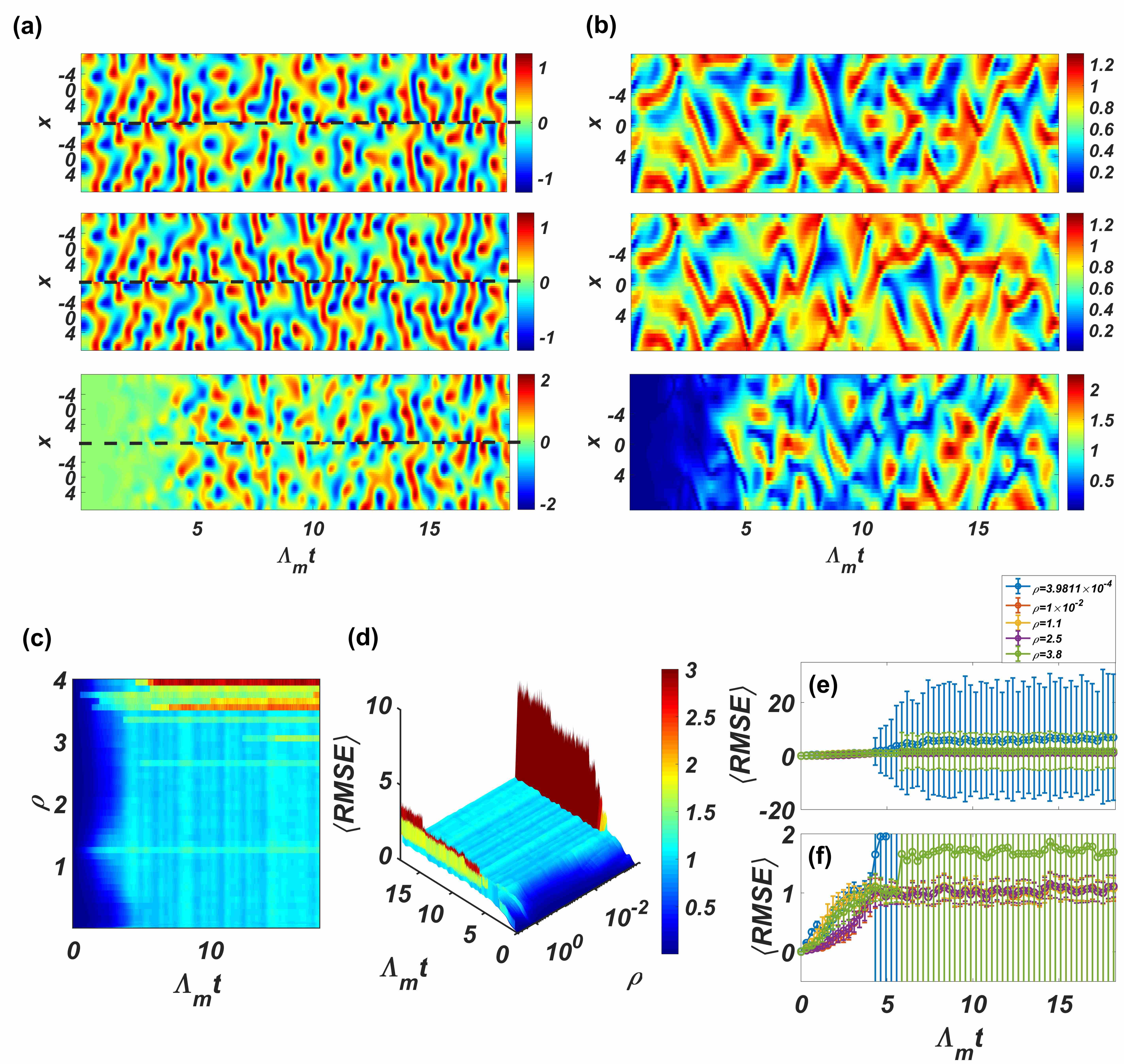}
\caption{ {\em Emergence of a valley interval in the spectral radius of the
reservoir network with minimum error for predicting spatiotemporal chaotic
solutions of the CGLE}. (a,b) An example of successful prediction of 
spatiotemporal chaotic solution of the one-dimensional CGLE, 
for which the maximum Lyapunov 
exponent is $\Lambda_m \approx 0.22$. In (a), the true and predicted real 
and imaginary parts (above and below the horizontal black dashed lines, 
respectively) of the spatiotemporal evolution of the solution, together with 
their difference, are shown. In (b), the true and predicted magnitude of 
the complex solution as well as their difference are displayed. The color bars 
in (a) indicate the scale of the real and imaginary parts of $u(x,t)$. For
both (a) and (b), the value of the spectral radius of the reservoir network 
is $\rho=0.1$. (c) Ensemble averaged RMSE calculated from the magnitude value 
of the complex solution versus $\rho$ and the Lyapunov time, where $100$ 
random reservoir systems are used for each fixed $\rho$ value. (d) A 
three-dimensional view of $\langle \mbox{RMSE}\rangle$, where the color bar 
indicates its scale with the cut-off value of $3.0$. The existence of a 
valley interval in $\rho$ that minimizes the prediction error is unequivocal. 
(e,f) Time evolution of $\langle \mbox{RMSE}\rangle$ (with standard deviation)
for five $\rho$ values. (Details of the statistical behavior of 
$\langle \mbox{RMSE}\rangle$ are presented in Appendix Secs.~\ref{app_sec:SD} 
and \ref{app_sec:SD_origin}. Other parameters are $\alpha=1$, $N=9984$, $k=3$, 
$M=L=64$, $N_{t}=80010$, $S=10$, and $\Gamma=2\times10^{-5}$.}
\label{fig:res_pre_CGL}
\end{figure*}

We test reservoir-computing based prediction of chaotic solutions of the KSE
(the original system that was used to demonstrate the power of reservoir 
computing to predict spatiotemporal chaotic systems~\cite{PHGLO:2018}) and
show the emergence of a valley interval in the spectral radius of the
reservoir network with minimum prediction error. The KSE is a one-dimensional 
nonlinear PDE given by
\begin{align} \label{eq:KSE}
\centering
\frac{\partial u}{\partial t} = - u \frac{\partial u}{\partial x}
- \frac{\partial^2 u}{\partial x^2} - \frac{\partial^4 u}{\partial x^4}
\end{align} 
where $u(x,t)$ is a scalar field. We set the one-dimensional spatial domain 
to be $x \in [0, 22]$. To obtain the true solution $u(x,t)$, we divide the 
spatial domain evenly using 64 grid points and numerically solve the KSE with 
time step $dt=0.25$. We thus have $64$ time series, one from each grid point.
Due to the chaotic nature of the KSE, even with reservoir computing it is not 
possible to predict the behavior of $u(x,t)$ for a relatively long time, and 
the demonstrated prediction horizon is a few Lyapunov time~\cite{PHGLO:2018}, 
defined as $\Lambda_m t$, with $\Lambda_m$ being the largest Lyapunov exponent 
of the chaotic solution. An example of successful prediction for about five 
Lyapunov time is shown in Fig.~\ref{fig:res_pre_KS}(a), where the value of 
RMSE is smaller than 0.5. Our main point is that, as for the case of NLSE, 
a valley interval in the spectral radius of the reservoir network with minimum
error emerges for the chaotic solution of the KSE, as illustrated in 
Figs.~\ref{fig:res_pre_KS}(b,c), where the interval is 
$0.02 \alt \rho \alt 0.25$. For $\rho < 0.02$, the ensemble averaged 
prediction error $\langle \mbox{RMSE}\rangle$ is significantly larger, as 
shown in Fig.~\ref{fig:res_pre_KS}(c). As the value of $\rho$ is increased 
from about 0.25, the prediction horizon reduces dramatically, as shown in 
Fig.~\ref{fig:res_pre_KS}(b). The time evolution behaviors of 
$\langle \mbox{RMSE}\rangle$ for four representative values of $\rho$ 
are shown in Fig.~\ref{fig:res_pre_KS}(d). We see that, for $\rho=0.0251$,
the errors are much larger than those for the other three cases. Thus, in spite
of the chaotic nature of the solution of the KSE, the valley phenomenon  
associated with reservoir computing based prediction still occurs, as 
for the regular solutions of the NLSE.

\subsection{Predicting spatiotemporal chaotic solutions of complex Ginzburg-Landau equation}

The CGLE is a general model for gaining insights into a variety of physical 
phenomena such as nonlinear waves, chemical reactions, superconductivity,
superfluidity, Bose-Einstein condensation, and liquid 
crystals~\cite{AK:2002,CH:1993,Kbook:1984}. The equation can generate 
solutions corresponding to complex physical phenomena such as phase chaos, 
defect chaos, coexistence of chaos and plane waves solution, etc. 
In a system described by the CGLE, instabilities lead to the formation of a 
weakly interacting and incoherent background of low-amplitude waves which, 
under certain conditions, can collapse locally to generate a large amplitude 
event. Because of this feature, the CGLE has been used in previous studies 
to characterize the statistical properties of the extreme events in 
spatiotemporal dynamical systems~\cite{KO:2003} and to articulate control 
strategies~\cite{NO:2007,DCLX:2008}.

In one spatial dimension, the CGLE is written as
\begin{align} \label{eq:CGLE}
\centering
\dfrac{\partial u}{\partial t}=u+(1+i\alpha)\frac{\partial^2 u}{\partial x^2}-(1+i\beta)|u|^2u,
\end{align}      
where $u(x,t)$ is a complex function of space $x$ and time $t$, $\alpha$ 
and $\beta$ are parameters characterizing linear and nonlinear dispersion, 
respectively. To be concrete, we focus on the parameter region of defect 
chaos~\cite{AK:2002}, e.g., $\alpha=2$ and $\beta=-2$. We set the spatial 
domain to be $x \in [-9, 9]$. To generate the data for the reservoir system
and take into consideration the dynamical complexity of the solutions of 
the CGLE, we solve Eq.~(\ref{eq:CGLE}) numerically using the pseudo-spectral
and exponential-time differencing scheme~\cite{CM:2002}, where the spatial
domain is divided uniformly into $32$ subregions and the integration time
step is $dt=0.0001$. From the numerical solutions, we perform time-domain
sampling with $dt=0.07$.

Because of the complex nature of the scalar field $u(x,t)$, two separate 
input-data streams to the reservoir system are necessary, corresponding to the 
real and imaginary parts of $u(x,t)$, respectively. (We have verified that the
reservoir system fails to produce any meaningful prediction if the module 
$|u(x,t)|$ is used as the input data.) Figures~\ref{fig:res_pre_CGL}(a,b)
show an example of successful prediction over a horizon of about four 
Lyapunov time, where the spectral radius value is $\rho = 0.1$. The existence
of an optimal valley interval in $\rho$ guaranteeing a similar prediction
performance is shown in Figs.~\ref{fig:res_pre_CGL}(c,d): 
$0.1 \alt \rho \alt 2.5$. When the value of $\rho$ is decreased from the 
left end of the interval, the ensemble averaged prediction error 
$\langle \mbox{RMSE}\rangle$ increases dramatically. Likewise, when
$\rho$ is increased from the right end of the interval (e.g., from $3.0$ 
to $4.0$), the predicted time with $\langle \mbox{RMSE}\rangle$ less than about
$0.5$ decreases monotonically, as shown in Fig.~\ref{fig:res_pre_CGL}(c). 
Figure~\ref{fig:res_pre_CGL}(e) presents the behaviors of the time evolution
of $\langle \mbox{RMSE}\rangle$ for five specific values of $\rho$. We 
see that the two cases where the values of the spectral radius are outside 
the valley interval (i.e., $\rho=3.9811\times10^{-4}$ and $\rho=3.8$),
large prediction errors arise. In fact, the standard deviation associated with 
the evolution is so large that a scale change in the vertical axis is 
necessary, as done in Fig.~\ref{fig:res_pre_CGL}(f). For other values of the
spectral radius, the standard deviation associated with 
$\langle \mbox{RMSE}\rangle$ is small when its value is less than about 0.5. 
When the value of $\langle \mbox{RMSE}\rangle$ becomes large and plateaued, the
values of the standard deviation are approximately uniform. The values of 
$\langle \mbox{RMSE}\rangle$ and its standard deviation for 
$\rho=1\times10^{-2}$ and $\rho=2.5$ are somewhat similar, but those for the 
case of $\rho=1.1$ are somewhat larger. In spite of the diverse behaviors of 
$\langle \mbox{RMSE}\rangle$ and its standard deviation, the valley phenomenon
giving rise to an optimal interval in the network spectral radius that 
minimizes the prediction error holds also for the 1D CGLE, indicating 
generality of the phenomenon.
  
\section{Effect Of reservoir network structure on prediction}

\begin{figure} [ht!] 
\centering
\includegraphics[width=\linewidth]{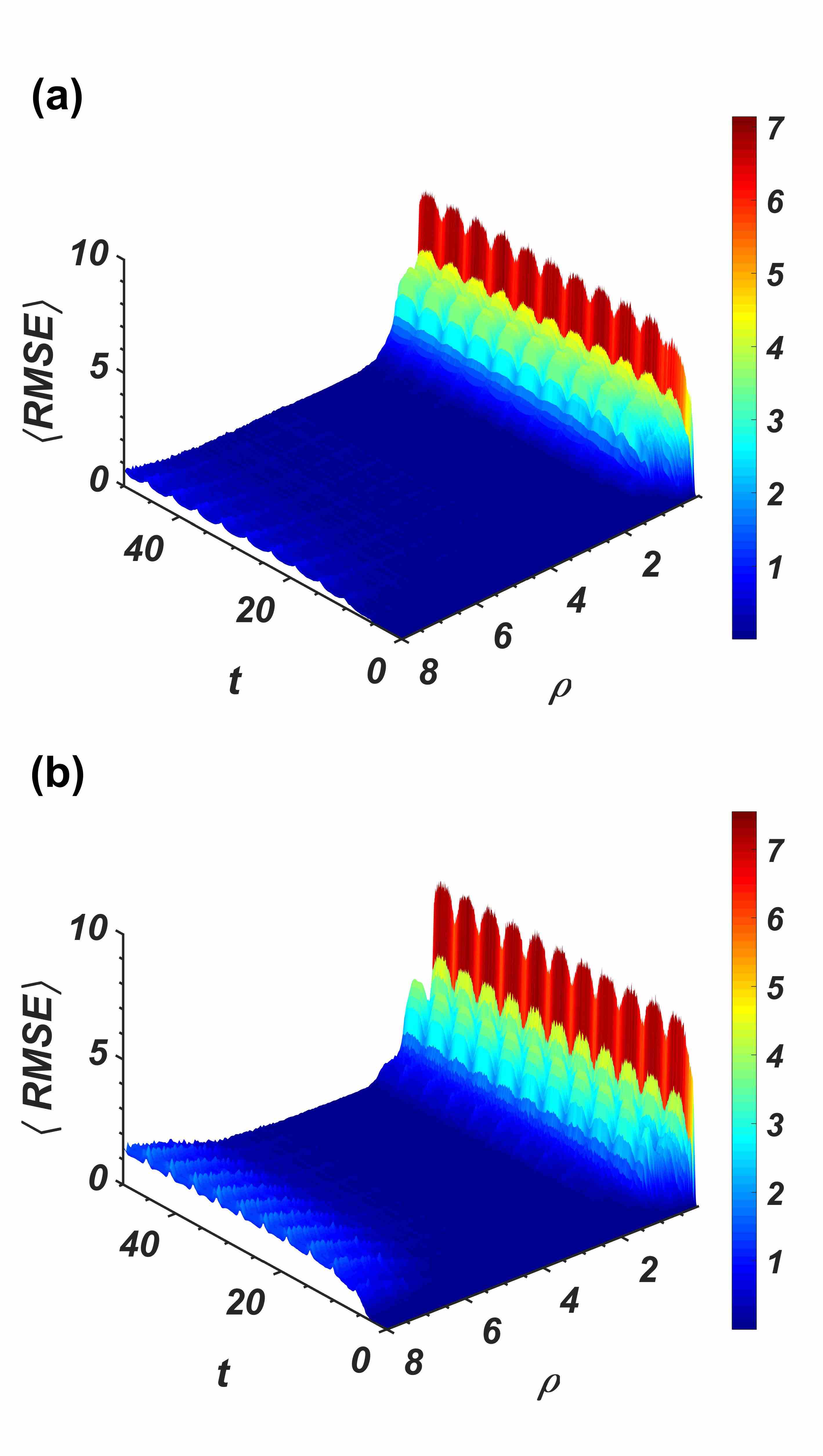}
\caption{ {\em Emergence of optimal valley interval in spectral radius with 
minimum prediction error for undirected random reservoir networks}. 
(a) For undirected random reservoir network, time evolution of the ensemble 
averaged RMSE (over $100$ random reservoirs) for systematically varying 
values of the network spectral radius for prediction of Akhmediev breathers
in the NLSE. (b) Similar plot but for undirected small-world reservoir networks,
where the value of the rewiring probability for generating the small-world
topology is 0.3. In both cases, a similar valley region arises, indicating that 
the link topology of the reservoir network, directed or undirected, has 
little effect on the emergence of the valley. The location and size of the
valley interval, however, do depend on the link topology, where undirected
networks tend to lead to a larger interval. Parameter values are $\alpha=1$, 
$N=4992$, $M=L=64$, $N_{t}=8010$, $S=10$, $\Gamma=1\times 10^{-4}$, $k=3$ 
for (a) and $k=4$ for (b).}
\label{fig:NS_BSRR}
\end{figure}

The random reservoir networks employed in the various examples in 
Sec.~\ref{sec:rho} all have directed edges. Will the existence of an optimal 
valley interval in spectral radius persist if the links in the reservoir
network become undirected? To address this question, we consider two types 
of undirected complex networks: random and small-world networks, and test
the prediction performance for Akhmediev breathers in the NLSE. 
Figures~\ref{fig:NS_BSRR}(a,b) show the time evolution of the ensemble 
averaged prediction error for systematically varying values of $\rho$
for undirected random and small-world networks, respectively. A quite sizable
valley interval in $\rho$ with minimum prediction error arises in each case.
In fact, in comparison with the directed network structure, the undirected 
topology leads to a wider valley interval [e.g., comparing 
Fig.~\ref{fig:NS_BSRR}(a) with Fig.~\ref{fig:res_pre_ABs}(c)].
A comparison between Fig.~\ref{fig:NS_BSRR}(a) and Fig.~\ref{fig:NS_BSRR}(b)
indicates that the valley interval for the random network structure is slightly
larger than that for the small-world topology. In general, whether the network
structure is directed or undirected not only affects the size of the valley
interval, but also leads to different ``best'' value of the spectral radius 
for which an absolute minimum in the prediction error can be achieved. 
We have tested other dynamical patterns in the NLSE as well as the KSE and
the CGLE and found that the existence of the best spectral radius region is 
robust, regardless of whether the edges in the reservoir network are 
directed or undirected.

\section{Error in training output data}

\begin{figure} [ht!] 
\centering
\includegraphics[width=\linewidth]{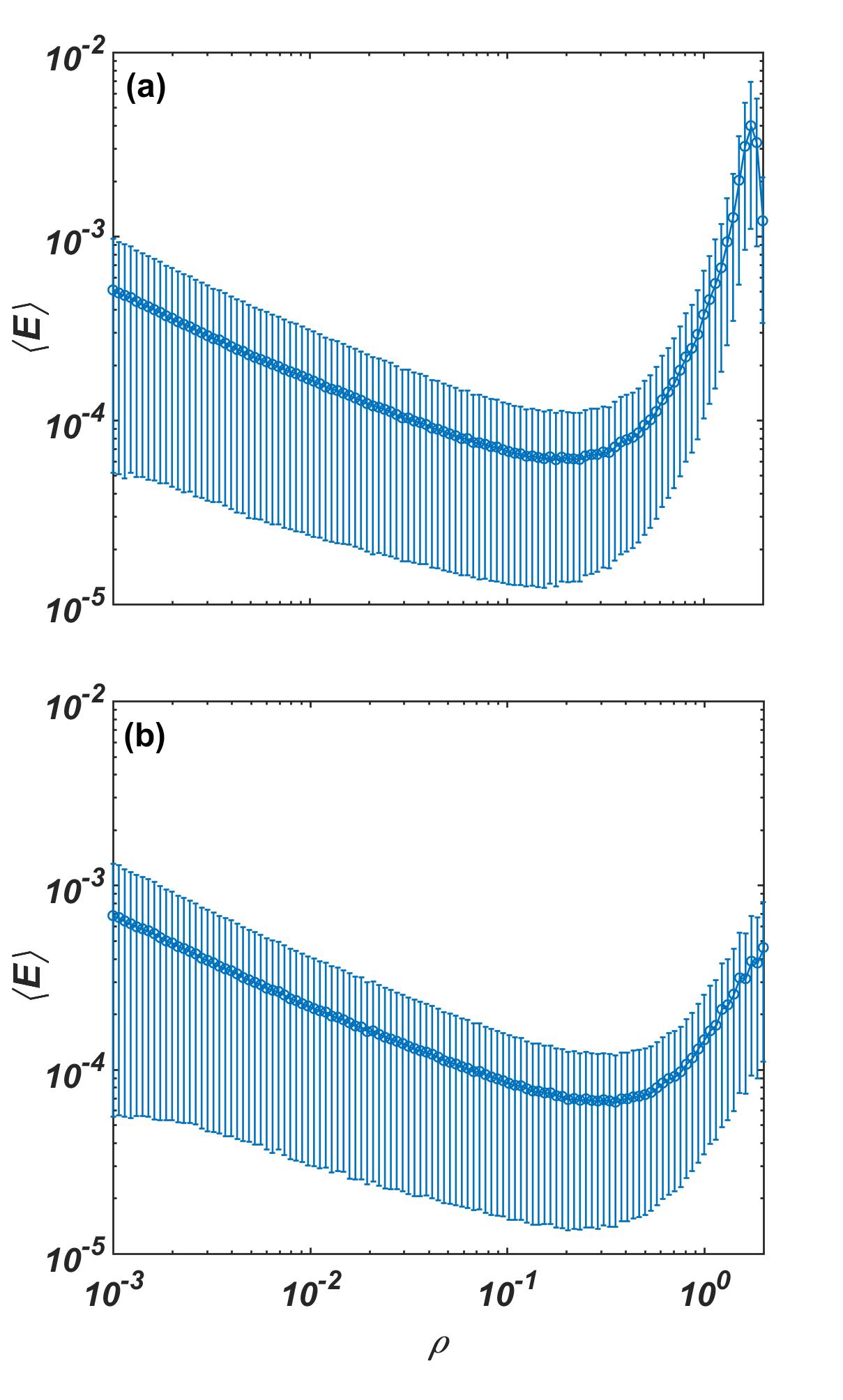}
\caption{ {\em Behavior of error during the training phase}. 
The time averaged error $\langle E\rangle$ versus the spectral radius $\rho$ 
for a random reservoir network with (a) a directed topology and (b) an 
undirected topology with the same average degree as in (a). In each case, the 
network structure is fixed but the link weights are adjusted to result in 
systematic variations in the network spectral radius. In both cases, a region
of small errors arises, indicating the existence of an optimal interval of 
spectral radius after training. The location and size of the region are 
similar to the valley interval in, e.g., Fig.~\ref{fig:NS_BSRR}. Parameter 
values are $\alpha=1$, $N=4992$, $k=3$, $M=L=64$, $N_{t}=70010$, $S=10$, and 
$\Gamma=1\times 10^{-4}$.}
\label{fig:ER_D_UD_KSE}
\end{figure}

To gain insights into the behavior of ensemble averaged RMSE in prediction, we 
examine the error associated with the training phase. 
From Eq.~(\ref{eq:res_evol}), we define the time averaged error during the 
training phase as $E=||\mathbf{W}_{RO}\cdot \mathbf{r}'-\mathbf{v}||$, which 
measures the difference between the generated and true training output state 
vector of the neural network, i.e., the error in one time step after
training. Figure~\ref{fig:ER_D_UD_KSE}(a) shows the time averaged error 
$\langle E\rangle$ versus the spectral radius $\rho$ for KSE with a directed 
network structure, which exhibits a non-monotonic behavior. Note that, the 
value of $\rho$ minimizing the error is within the valley 
interval in Fig.~\ref{fig:res_pre_KS}(b). The increase in the error away 
from the minimum value as $\rho$ is increased corresponds to the decrease in 
the prediction horizon in Fig.~\ref{fig:res_pre_KS}(b). However, the behavior 
of $\langle E\rangle$ as $\rho$ is decreased from the optimal value does not 
appear to explain the dramatic increase in the ensemble averaged RMSE in 
prediction in Fig.~\ref{fig:res_pre_KS}(c). Figure~\ref{fig:ER_D_UD_KSE}(b) 
shows a similar behavior of $\langle E\rangle$ but for the case where the 
complex neural network has an undirected topology. At the present, the 
behaviors of error growth on the two sides of the valley have not been 
analytically understood.

Figure~\ref{fig:ER_D_UD_KSE} offers insights into the source of prediction 
error with implications to the prediction time that reservoir computing 
can possibly achieve. From Fig.~\ref{fig:ER_D_UD_KSE}, we see that the 
smallest average predicting error for each step is about
$6\times10^{-5}$ for KSE. For the spatiotemporal chaotic solution of KSE, 
the maximum Lyapunov exponent is $\Lambda_m\approx 0.05$. With time step 
$dt=0.25$, in the predicting phase, the error will grow to about 0.5 in 
five Lyapunov time. The prediction time is thus mainly determined by the 
prediction error of reservoir computing at each time step. While the 
single-step prediction error can be reduced to certain extent by fine-tuning 
the parameters of the neural network, such reduction is often incremental and
there is no general method at the present to drastically reduce the          
single-step error.

\section{Discussion} \label{sec:discussion}

Reservoir computing, a class of recurrent neural networks articulated nearly 
two decades ago~\cite{MNM:2002,MJ:2013,JH:2004} for data-based prediction 
of nonlinear dynamical systems, has recently gained momentum~\cite{HSRFG:2015,
LBMUCJ:2017,PLHGO:2017,LPHGBO:2017,PWFCHGO:2018,PHGLO:2018,Carroll:2018,
NS:2018,ZP:2018,WYGZS:2019} as stimulated by the significant growth of 
interest and tremendous advances in modern machine learning. For chaotic 
dynamical systems, traditional methods~\cite{FS:1987,Casdagli:1989,BT:1992,
Petrov:1997} based on delay-coordinate embedding~\cite{Takens:1980} can
usually make short-term prediction, e.g., for about one Lyapunov time.
Another prediction framework is based on sparse 
optimization such as compressive sensing~\cite{WYLKG:2011,WLG:2016}, but this 
approach requires that the system's equations contain mathematically simple 
terms and time series data from all variables of the system be available.
Reservoir-computing based prediction is model free and solely data based, and
it can extend the horizon to about half dozen Lyapunov time.
This is quite remarkable, defying the conventional 
wisdom that long-term prediction of the state evolution of a chaotic system 
is ruled out due to the hallmark of chaos: sensitive dependence on initial 
conditions. A reservoir computing system, fundamentally being a large neural
network, has a large number of parameters whose values need to be fixed. 
While the values of a subset of parameters can be determined through training 
with available data, there are still many ``free'' parameters whose values
need to be pre-set. At the present, for reservoir computing (or for machine 
learning), there are no general rules that one can rely on to guide the 
choices of these parameters. Due to the vast complexity and nonlinear 
structure of reservoir computing systems, to develop mathematical or physical
theories to guide systematic choices of the values of free parameters 
represents an outstanding and formidably challenging problem, with no hope for
solutions in sight.  

To make progress, we focus on a spectral property of the reservoir network 
that typically possesses a complex topology (e.g., random or small-world): 
the spectral radius. Such a network is typically weighted with heterogeneous
weights distributed on the set of edges. With variations in the detailed 
connecting topology and link weights, for the network alone, the parameter 
space is vast. To make the exploration feasible, we fix the connection 
topology and assume that only the link weights can vary freely. Even then, 
combing through all possible parameter variations is a computationally 
prohibitive task. We thus focus on one question: is there a range of the 
spectral radius value that can lead to optimal performance in the sense of 
minimum prediction error? Note that, with a fixed value of the spectral 
radius, there are still an infinite number of sets of link weights. 
Computations with three representative classes of spatiotemporal nonlinear 
dynamical systems (the NLSE, the KSE, and the CGLE) reveal a remarkable 
phenomenon: in all cases there exists an optimal interval in the spectral 
radius that leads to minimum error. (In the three-dimensional plot of the 
ensemble averaged prediction error versus spectral radius and time, the 
interval appears as a ``valley''.) The existence of such a valley interval
holds generally true for different systems, regardless of the structure of the 
complex reservoir network, e.g., directed or undirected, random or small-world.
Computationally, we find that the interval tends to be larger for undirected
than for directed networks. While the finding is purely numerical with no
analytic insights, the phenomenon is general and can be exploited for 
designing optimal reservoir computing systems, representing a small step 
forward in the study of these machine learning systems.

At the present, we do not yet have an analytic understanding as to why the
value of the spectral radius $\rho$ of the reservoir network needs to be in a
certain interval for the neural network system to be effective for prediction.
Nonetheless, a heuristic understanding may be attempted. In
order for the reservoir system to possess certain predictive power, it must
capture the ``dynamical climate or complexity'' of the target nonlinear
system through training. That is, the reservoir system must produce state
evolution whose complexity somehow matches that of the target system. In our
setting, the network topology is fixed and the variations in the spectral
radius are the result of adjusting the edge weights. If the spectral radius is
too small, the edge weights are small and the network may be so weakly
connected that its collective dynamics are too incoherent to match that of
the target system. However, if the spectral radius is too large, the nodal
connections in the reservoir network are so tight that the collective dynamics
are too coherent, depriving the reservoir computing system of its ability
to capture the ``climate'' of the state evolution of the target system.
As a balance of these factors, it is reasonable that, given training data from
the dynamical evolution of a specific target system, in general
an interval in $\rho$ should emerge in which an optimal match between the
complexity of the two systems is achieved. The particular location and
size of the interval would depend on the details of the target system to be
predicted.

Our work has raised more open questions. For example, a previous work 
demonstrated that the echo state property of reservoir computing can be 
ensured for $\rho < 1$ with zero input but, for non-zero input, the value of 
$\rho$ can be extended to being larger than one~\cite{LJ:2009}. Our study has 
revealed that, for both the NLSE and the 1D CGLE, the optimal interval in the 
spectral radius is located in the region $\rho > 1$. Another previous 
speculation was to regard the spectral radius as a kind of measure of 
reservoir's memory length of the input signal. Consequently, if the input 
signals are more random and require a larger memory to store, one should 
employ a reservoir network with a larger spectral radius for 
prediction~\cite{LJ:2009,lukovsevivcius:2012}. However, our results do not 
support this point of view. For example, for the NLSE, the dynamical patterns 
studied are periodic either in space or in time and are thus mostly regular 
with a minimum degree of randomness, and yet the optimal valley intervals of 
$\rho$ can be quite different. Why patterns of similar regularity require 
different spectral-radius values to be predicted? For the CGLE, in spite of 
the randomness and complexity of the its dynamical evolution, the valley 
interval is relatively more extensive from near zero values to some values 
far beyond one. Why can the quite random and complex patterns of the CGLE 
be predicted with reservoirs of either long or short memory capacity?

\begin{acknowledgments}

This work is supported by the Pentagon Vannevar Bush Faculty Fellowship
program sponsored by the Basic Research Office of the Assistant Secretary of
Defense for Research and Engineering and funded by the Office of Naval
Research through Grant No.~N00014-16-1-2828.

\end{acknowledgments}

\begin{appendix}

\section{Standard deviation of prediction error} \label{app_sec:SD}

\begin{figure} [ht!]
\centering
\includegraphics[width=\linewidth]{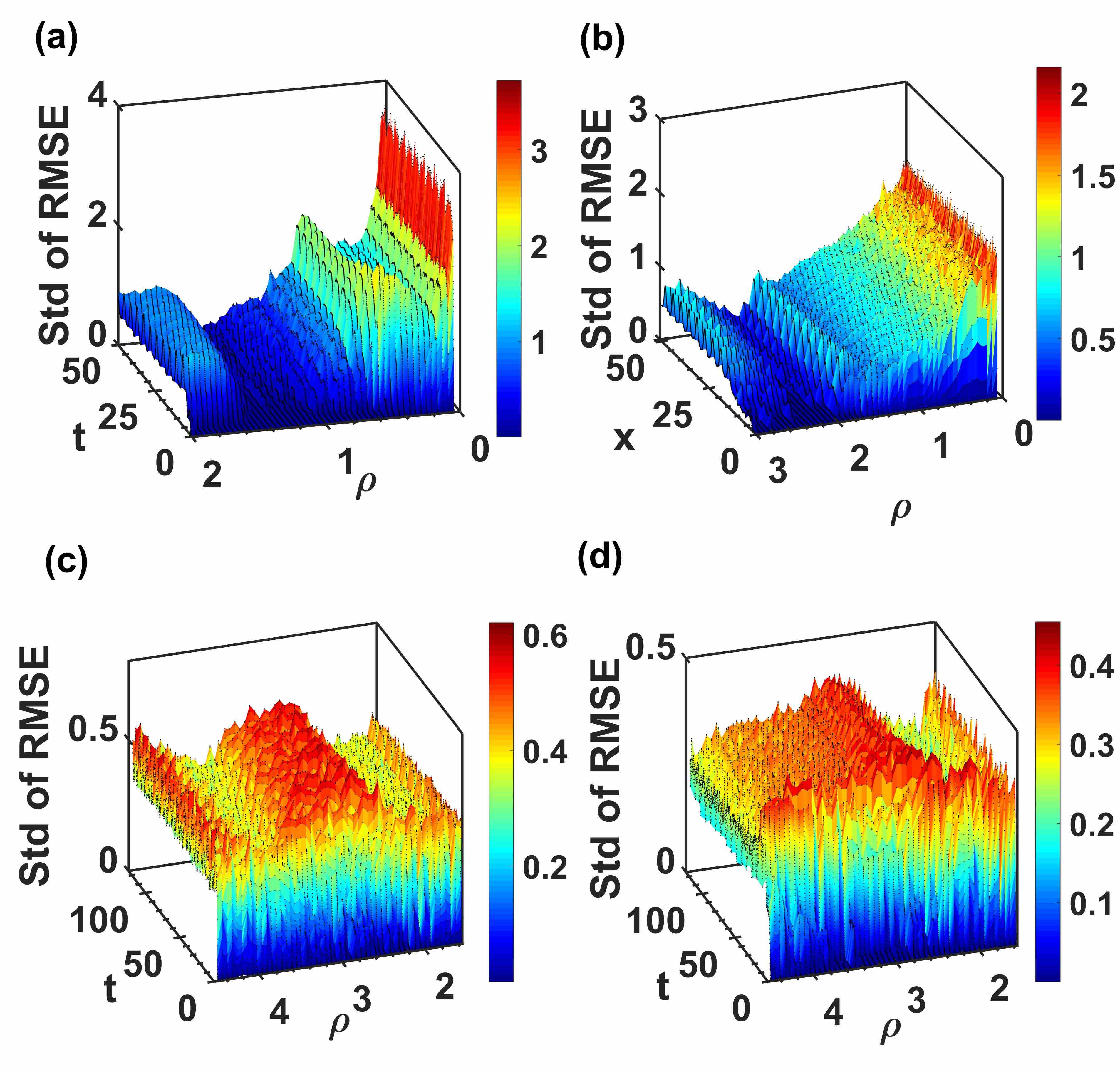}
\caption{ {\em Standard deviation associated with ensemble-averaged 
error in predicting different dynamical states of the NLSE.}
For each fixed value of the spectral radius $\rho$,
100 random realizations of the network are used to calculate the
standard deviation of the ensemble-averaged prediction error. Shown are the 3D
representation of the standard deviation versus time and $\rho$ for four
distinct dynamical states of the NLSE: (a) Akhmediev breathers,
(b) Kuznetsov-Ma solitons, (c) a soliton-collision state for $a_1=0.14$ and
$a_2=0.34$, and (d) another soliton-collision state for $a_1=0.42$ and
$a_2=0.18$.}
\label{fig:ABs_std}
\end{figure}

In the main text, we have presented the ensemble-averaged error
$\langle \mbox{RMSE}\rangle$ versus the spectral radius of the reservoir
network and time, which arises from predicting various states of three types 
of spatiotemporal dynamical systems. Here we show the standard deviation 
associated with the error, as in Figs.~\ref{fig:ABs_std}(a-d) for the 
corresponding cases. In particular, in Figs.~\ref{fig:ABs_std}(a,b) where the 
target states are Akhmediev breathers and Kuznetsov-Ma solitons, respectively, 
the values of the standard deviation are small in the valley region but 
increase as the value of $\rho$ moves out of the valley, indicating that 
stable prediction performance can be achieved when choosing the value of 
$\rho$ in the valley. In Figs.~\ref{fig:ABs_std}(c,d) where the dynamical 
states are two distinct cases of soliton collision, the standard deviations 
is large for all values of $\rho$ tested.

\begin{figure} [ht!]
\centering
\includegraphics[width=\linewidth]{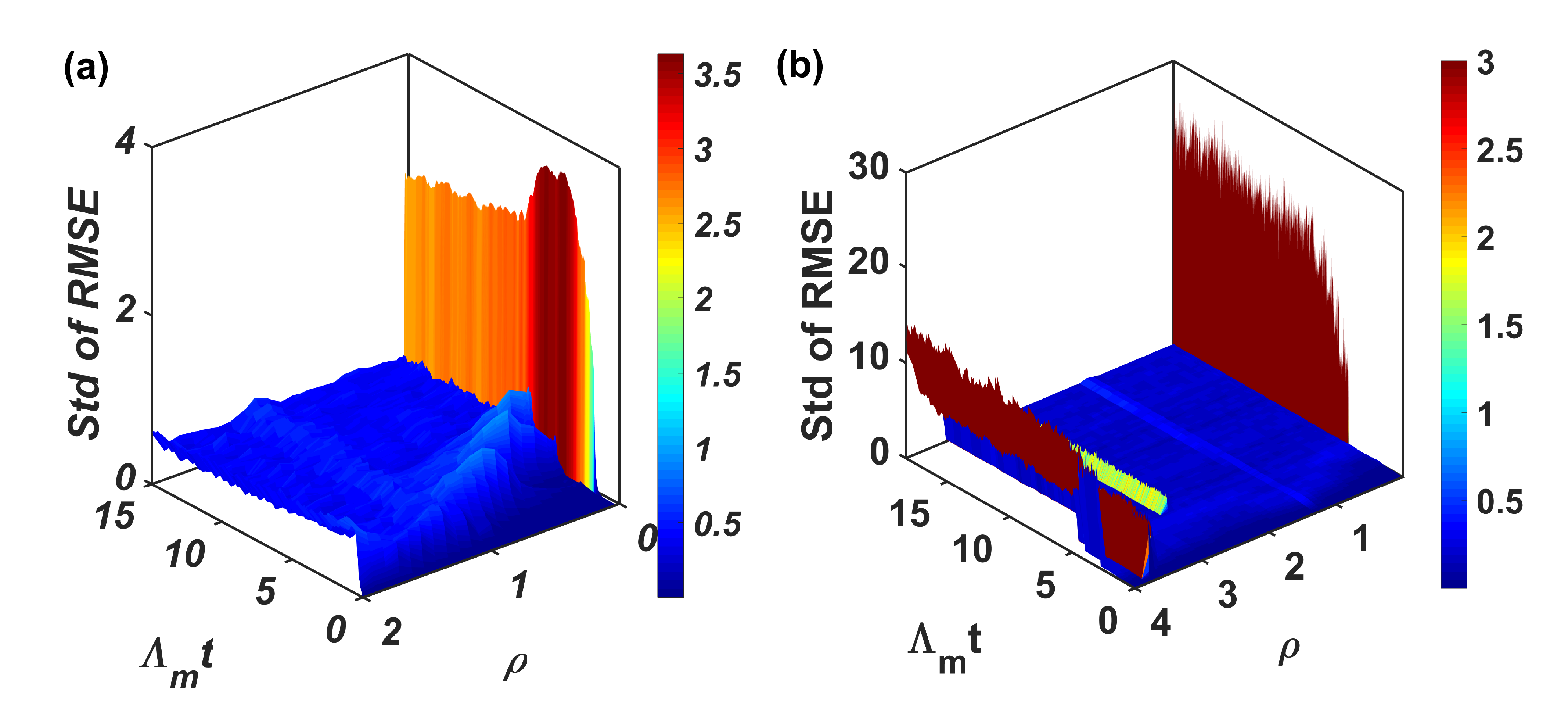}
\caption{ {\em Standard deviation associated with ensemble-averaged 
prediction error for the KSE (a) and 1D CGLE (b).}
Legends are the same as in Fig.~\ref{fig:ABs_std}.}
\label{fig:KS_CGL_std}
\end{figure}

Results of the standard deviation for the KSE and 1D CGLE are shown 
in Fig.~\ref{fig:KS_CGL_std}, where the dynamical states to be predicted are
spatiotemporally chaotic. Again, we observe that the standard deviation
associated with the ensemble-averaged error is markedly smaller in the
valley interval in the network spectral radius than outside the interval.

\section{Example of long term prediction of Akhmediev breathers in NLSE}
\label{app_sec:LTP}

\begin{figure} [ht!]
\centering
\includegraphics[width=\linewidth]{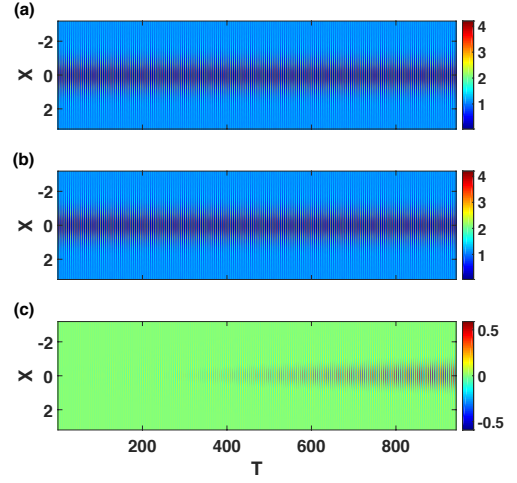}
\caption{ {\em An example of long-term prediction of Akhmediev breathers in
the NLSE}. (a) The true spatiotemporal evolution pattern, (b) the
reservoir-computing predicted pattern, and (c) the difference in the
instantaneous state between the true and predicted patterns.}
\label{fig:ABs_LT_Pre}
\end{figure}

For the dynamical state of Akhmediev breather in the NLSE, for properly chosen
values of the spectral radius, the reservoir computing systems is able to make
accurate long-term prediction. An example is shown in Fig.~\ref{fig:ABs_LT_Pre}.

\section{Origin of standard deviation in the ensemble-averaged prediction error}
\label{app_sec:SD_origin}

\begin{figure} [ht!]
\centering
\includegraphics[width=\linewidth]{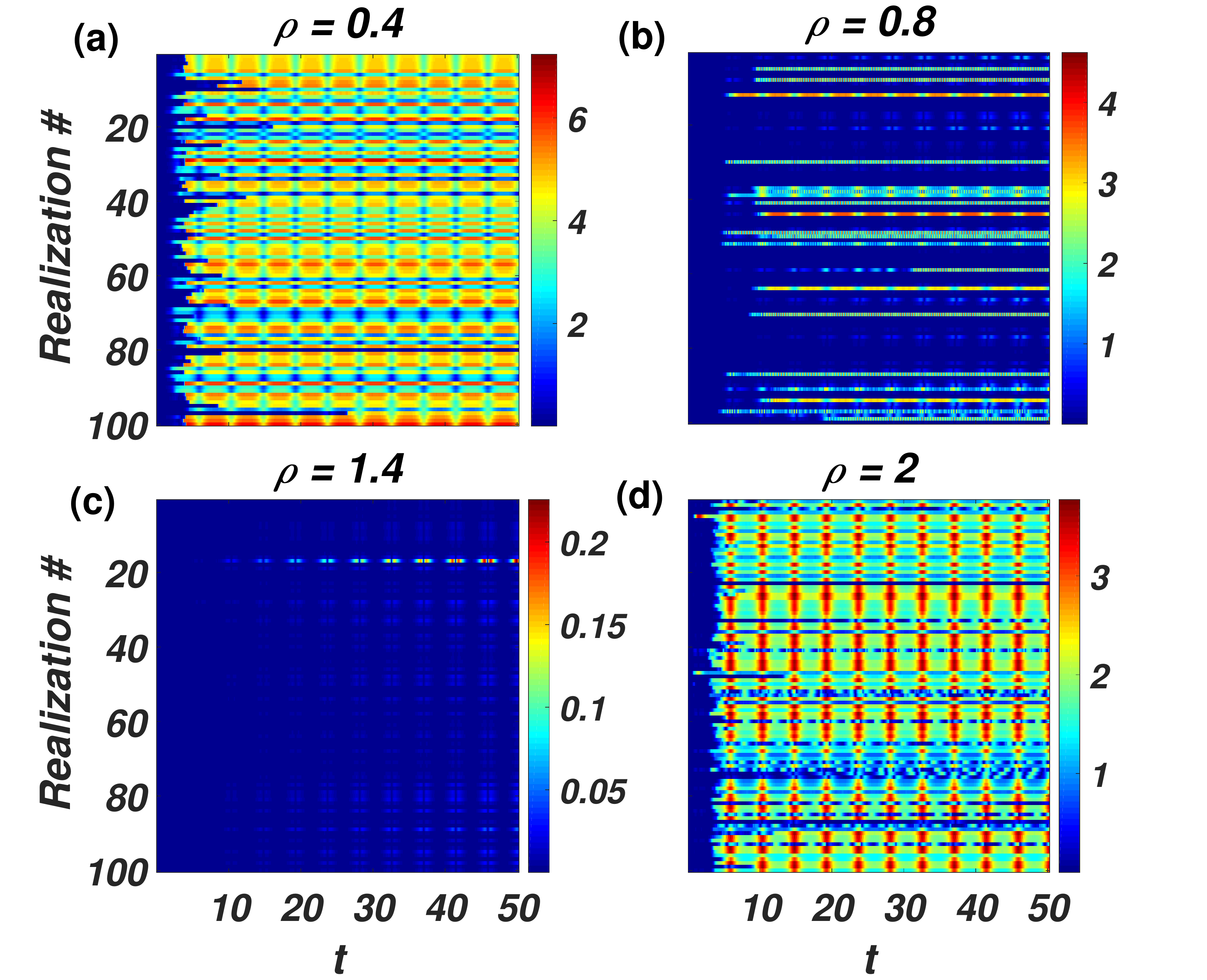}
\caption{ {\em Time evolution of RMSE for different statistical
realizations in predicting Akhmediev breathers in the NLSE}. Four cases
are shown, each for a fixed $\rho$ value. For $\rho$ inside the valley
interval, most or all realizations exhibit small errors, as in (b,c).
For $\rho$ outside the interval, almost all realizations exhibit large
errors, as in (a,d).}
\label{fig:ABs_100_reali}
\end{figure}

The concept of valley interval discussed in the main text is defined with
respect to the ensemble-averaged prediction error. That is, for any fixed
value of the spectral radius, 100 realizations of the reservoir network
are used to calculate the mean error and the standard deviation. In fact, over
the different realizations, the prediction error can exhibit quite large
variations, even when the value of the spectral radius is 
inside the valley. Several examples for predicting the Akhmediev breathers in
the NLSE are shown in Fig.~\ref{fig:ABs_100_reali}, where error evolution
for different realizations (ordinate) is shown for four different values of
the spectral radius (a-d). For the two cases outside the valley interval
(a,d), the prediction error is large across almost all the realizations. For
(c) $\rho = 1.4$, the error is small for almost all the realizations,
corresponding to the optimal $\rho$ value in the valley. When $\rho$ deviates
from the optimal value, large errors arise with some realizations, as shown
in (b) for $\rho = 0.8$. When majority of the realizations exhibit large
errors, the corresponding $\rho$ value is regarded as being outside the
valley interval. The error variations across different realizations are
characterized by the standard deviation in the mean error. For the optimal
$\rho$ value, the standard deviation reaches minimum. For $\rho$ away from the
optimal value, the standard deviation tends to increase. We also note that
the concept of valley interval is meaningful only in an approximate sense:
neither the ensemble-averaged error nor the associated standard deviation
presents any abrupt changes that can be used to define sharp boundaries of
the valley interval.

\begin{figure} [ht!]
\centering
\includegraphics[width=\linewidth]{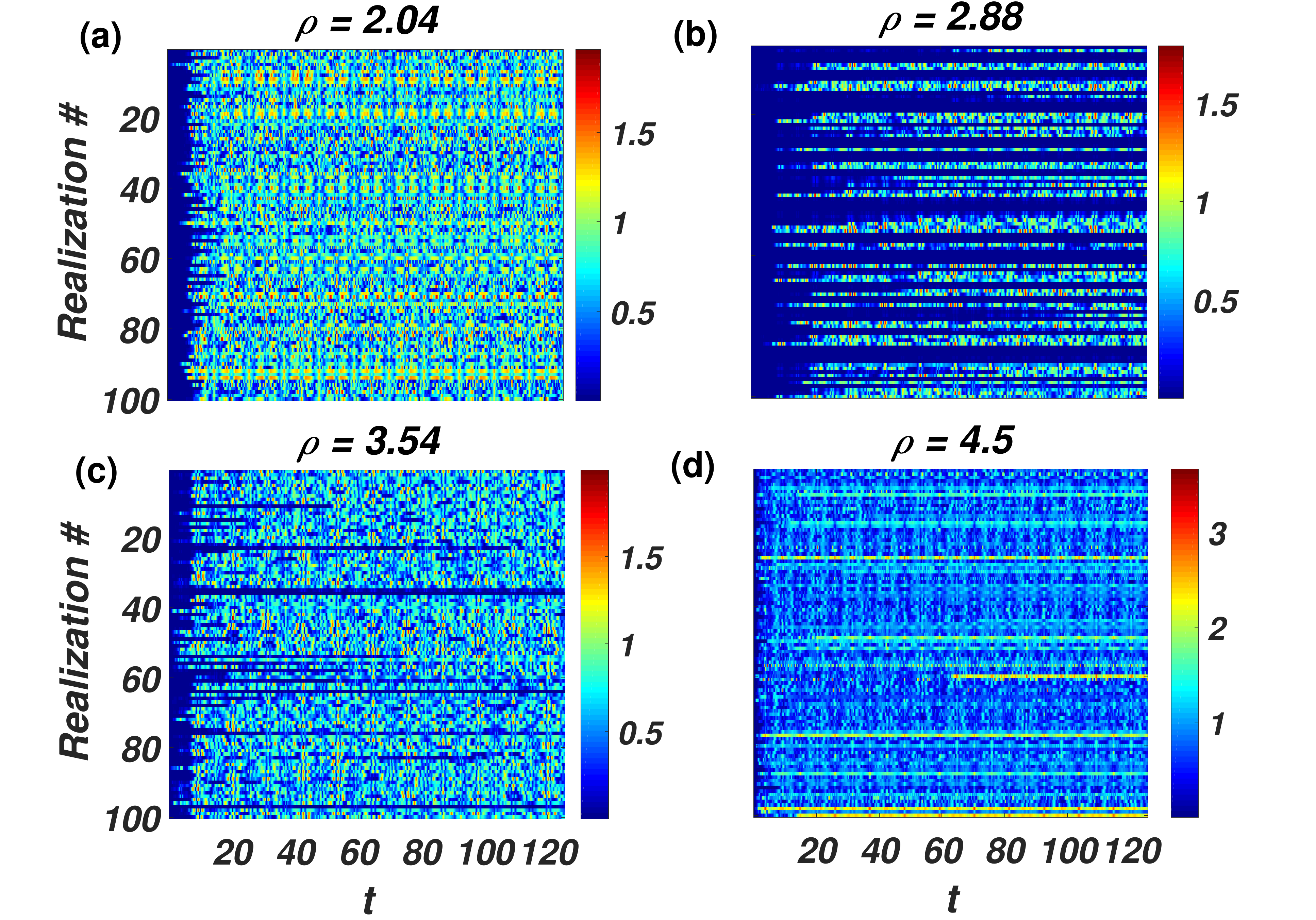}
\caption{ {\em Time evolution of RMSE for different statistical
realizations in predicting soliton collisions in the NLSE}. The parameters
of the NLSE solution are $a_1=0.14$, $a_2=0.34$. Legends are the same as
in Fig.~\ref{fig:ABs_100_reali}.}
\label{fig:ABs_coll_100_reali}
\end{figure}

\begin{figure} [ht!]
\centering
\includegraphics[width=\linewidth]{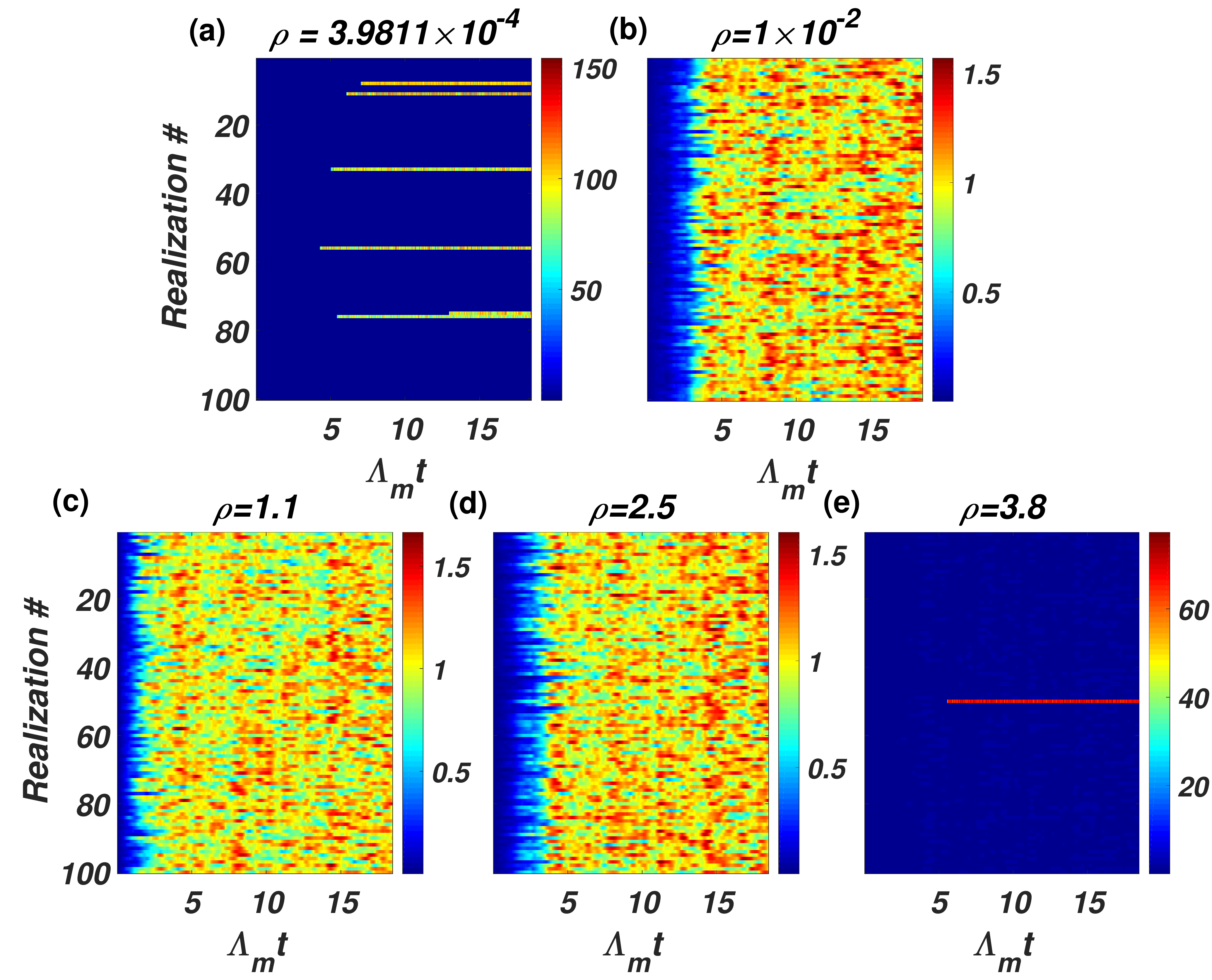}
\caption{ {\em Time evolution of RMSE for different statistical
realizations in predicting the spatiotemporal chaotic state of the
1D CGLE}. The parameter values of the CGLE are the same as those in
Fig.~5 in the main text. Legends are the same as those in
Fig.~\ref{fig:ABs_100_reali}.}
\label{fig:1DCGL_100_reali}
\end{figure}

The variations of the prediction error across individual realizations
for the soliton-collision state in the NLSE are shown in
Fig.~\ref{fig:ABs_coll_100_reali}, and the corresponding behaviors for
predicting the spatiotemporal chaotic state of the 1D CGLE are shown in
Fig.~\ref{fig:1DCGL_100_reali}.

\section{Solution of soliton collision in NLSE} \label{app_sec:NLSE_solution}

The complete solution of soliton collision in the NLSE is
given by~\cite{ASA:2009,FKM:2013}
\begin{widetext}
\begin{align}
\centering
\psi_{12}(x,t)&=\psi_0+\frac{2(l_1^*-l)s_1r_1^*}{|r_1|^2+|s_1|^2}+\frac{2(l_2^*-l_2)s_{12}r_{12}^*}{|r_{12}|^2+|s_{12}|^2},\\
r_1(x,t)&=\exp(\dfrac{-ix}{2})\bigg[\exp(\dfrac{i(2\chi_1+\kappa_1t-\pi/2+l_1\kappa_1x)}{2})-\exp(\dfrac{i(-2\chi_1-\kappa_1t+\pi/2-l_1\kappa_1x)}{2})\bigg],\\
s_1(x,t)&=\exp(\dfrac{ix}{2})\bigg[\exp(\dfrac{i(-2\chi_1+\kappa_1t-\pi/2+l_1\kappa_1x)}{2})+\exp(\dfrac{i(2\chi_1-\kappa_1t+\pi/2-l_1\kappa_1x)}{2})\bigg],\\
r_2(x,t)&=\exp(\dfrac{-ix}{2})\bigg[\exp(\dfrac{i(2\chi_2+\kappa_2t-\pi/2+l_2\kappa_2x)}{2})-\exp(\dfrac{i(-2\chi_2-\kappa_2t+\pi/2-l_2\kappa_2x)}{2})\bigg],\\
s_2(x,t)&=\exp(\dfrac{ix}{2})\bigg[\exp(\dfrac{i(-2\chi_2+\kappa_2t-\pi/2+l_2\kappa_2x)}{2})+\exp(\dfrac{i(2\chi_2-\kappa_2t+\pi/2-l_2\kappa_2x)}{2})\bigg],\\
r_{12}(x,t)&=\dfrac{(l_1^*-l_1)s_1^*r_1s_2+(l_2-l_1)|r_1|^2r_2+(l_2-l_1^*)|s_1|^2r_2}{|r_1|^2+|s_1|^2},\\
s_{12}(x,t)&=\dfrac{(l_1^*-l_1)s_1r_1^*r_2+(l_2-l_1)|s_1|^2s_2+(l_2-l_1^*)|r_1|^2s_2}{|r_1|^2+|s_1|^2},\\
\psi_0(x,t)&=\exp(ix),
\end{align}
\end{widetext}
where $l_1=i\sqrt{2a_1}$, $l_2=i\sqrt{2a_2}$, $\kappa_1=2\sqrt{1+l_1^2}$,
$\kappa_2=2\sqrt{1+l_2^2}$, $\chi_1=\frac{1}{2}\arccos(\kappa_1/2)$,
$\chi_2=\frac{1}{2}\arccos(\kappa_2/2)$, and $*$ represents the complex
conjugate.

\end{appendix}

\clearpage

%
\end{document}